\documentclass{article}

\PassOptionsToPackage{numbers, compress}{natbib}
\usepackage[final]{nips_2016}

\usepackage[utf8]{inputenc} % allow utf-8 input
\usepackage[T1]{fontenc}    % use 8-bit T1 fonts
\usepackage{hyperref}       % hyperlinks
\usepackage{url}            % simple URL typesetting
\usepackage{booktabs}       % professional-quality tables
\usepackage{amsfonts}       % blackboard math symbols
\usepackage{nicefrac}       % compact symbols for 1/2, etc.
\usepackage{microtype}      % microtypography

\usepackage{amsthm,amsmath,amssymb}
\usepackage{macros}
\usepackage{subcaption}
\usepackage[textfont=small, labelfont=small]{caption}
\usepackage{graphicx}
\DeclareGraphicsExtensions{.pdf,.png,.jpg,.eps}

\usepackage{algorithm}
\usepackage{algorithmic}

\usepackage[color=yellow]{todonotes}
\usepackage{booktabs}
\usepackage[inline]{enumitem}
\usepackage{verbatim}

\title{Bayesian latent structure discovery from multi-neuron recordings}

\author{Scott W. Linderman \\
  Columbia University \\
  \texttt{swl2133@columbia.edu} \\
  \And
  Ryan P. Adams \\
  Harvard and Twitter Cortex\\
  \texttt{rpa@seas.harvard.edu}\\
  \And
  Jonathan W. Pillow \\
  Princeton University \\
  \texttt{pillow@princeton.edu}
}

\begin{document}

\maketitle

\begin{abstract}
  Neural circuits contain heterogeneous groups of neurons that differ
  in type, location, connectivity, and basic response
  properties. However, traditional methods for dimensionality
  reduction and clustering are ill-suited to recovering the structure
  underlying the organization of neural circuits. In particular, they
  do not take advantage of the rich temporal dependencies in
  multi-neuron recordings and fail to account for the noise in neural
  spike trains. Here we describe new tools for inferring latent
  structure from simultaneously recorded spike train data using a
  hierarchical extension of a multi-neuron point process model
  commonly known as the generalized linear model (GLM). Our approach
  combines the GLM with flexible graph-theoretic priors governing the
  relationship between latent features and neural connectivity
  patterns.
  %allowing for Bernoulli, binomial (``under-dispersed''), and negative-binomial (``over-dispersed'')
  %spiking variability. 
  Fully Bayesian inference via \polyagamma
  augmentation of the resulting model allows us to classify neurons
  and infer latent dimensions of circuit organization from correlated
  spike trains.  We demonstrate the effectiveness of our method with
  applications to synthetic data and multi-neuron recordings in primate retina,
  revealing latent patterns of neural types and locations from spike trains alone.
% The resulting methodology will be broadly
%   useful for identifying network structure and functional
%   organization, particularly in the many brain areas where traditional
%   methods have failed to reveal coherent organizing principles.
\end{abstract}

\section{Introduction}
Large-scale recording technologies are revolutionizing the field of
neuroscience~\citep[e.g.,][]{ahrens2013whole, grewe2010high, prevedel2014simultaneous}.
These advances present an unprecedented opportunity to probe
the underpinnings of neural computation, but they also pose an
extraordinary statistical and computational challenge: how do we make
sense of these complex recordings? 
To address this challenge, we need methods that not only capture variability in
neural activity and make accurate predictions, but also expose meaningful
 structure that may lead to novel hypotheses and
interpretations of the circuits under study. In short, we
need exploratory methods that yield interpretable
representations of large-scale neural data.

For example, consider a population of distinct retinal ganglion cells (RGCs).
These cells only respond to light within their small receptive field.
Moreover, decades of painstaking work have revealed a
plethora of RGC types \citep{sanes2015types}. Thus, it is natural to characterize
these cells in terms of their type and the location of their receptive 
field center. Rather than manually searching for such a representation 
by probing with different visual stimuli, here we develop a method to automatically discover this structure from correlated patterns of neural activity.

Our approach combines latent variable network models \citep{Hoff2008,
  orbanz2015bayesian} with generalized linear models of neural spike
trains \citep{Paninski-2004, Truccolo-2005, Pillow-2008, vidne2012modeling} in a
hierarchical Bayesian framework. The network serves as a bridge,
connecting interpretable latent features of interest to the temporal
dynamics of neural spike trains. Unlike many previous studies
\citep[e.g.,][]{brillinger1976identification, Gerhard-2013,
  soudry2013shotgun}, our goal here is not necessarily to recover true
synaptic connectivity, nor is our primary emphasis on prediction.
Instead, our aim is to explore and compare latent patterns of
functional organization, integrating over possible networks. To do so,
we develop an efficient Markov chain Monte Carlo (MCMC) inference
algorithm by leveraging \polyagamma augmentation to derive collapsed
Gibbs updates for the network. We illustrate the robustness and
scalability of our algorithm with synthetic data examples, and we
demonstrate the scientific potential of our approach with an
application to retinal ganglion cell recordings, where we recover the
true underlying cell types and locations from spike trains alone,
without reference to the stimulus.

%Many neural spike train analyses have overlooked the enormous
%representational advantages of networks. Great emphasis has been
%placed on recovering an underlying synaptic network from spike train
%observations \citep[e.g.][]{brillinger1976identification,
%  fletcher2011neural, Gerhard-2013, soudry2013shotgun}, adhering to a
%strict biophysical interpretation of the edges. In contrast,
%generalized linear models \citep{Paninski-2004, Truccolo-2005} have
%leveraged networks to predict future spiking activity
%\citep[e.g.][]{Pillow-2008}, treating the network as simply a
%statistical parameter. Here, however, we view the network as a bridge
%between interpretable latent variables of interest --- variables that
%shed light on the function of neural circuits --- and the complex
%dynamics of neural spike trains. We develop a Bayesian framework that
%combines structured network priors with generalized linear models of
%neural activity to explore intuitive representations of neural
%populations.  We illustrate our approach with an application to
%retinal ganglion cell recordings, and we demonstrate the scalability
%of our algorithm, as well as its advantages over alternative
%techniques, with a synthetic data example.

\begin{figure}[t]
  \centering
  \begin{subfigure}[b]{5in}
     \centering
     \includegraphics[width=\textwidth]{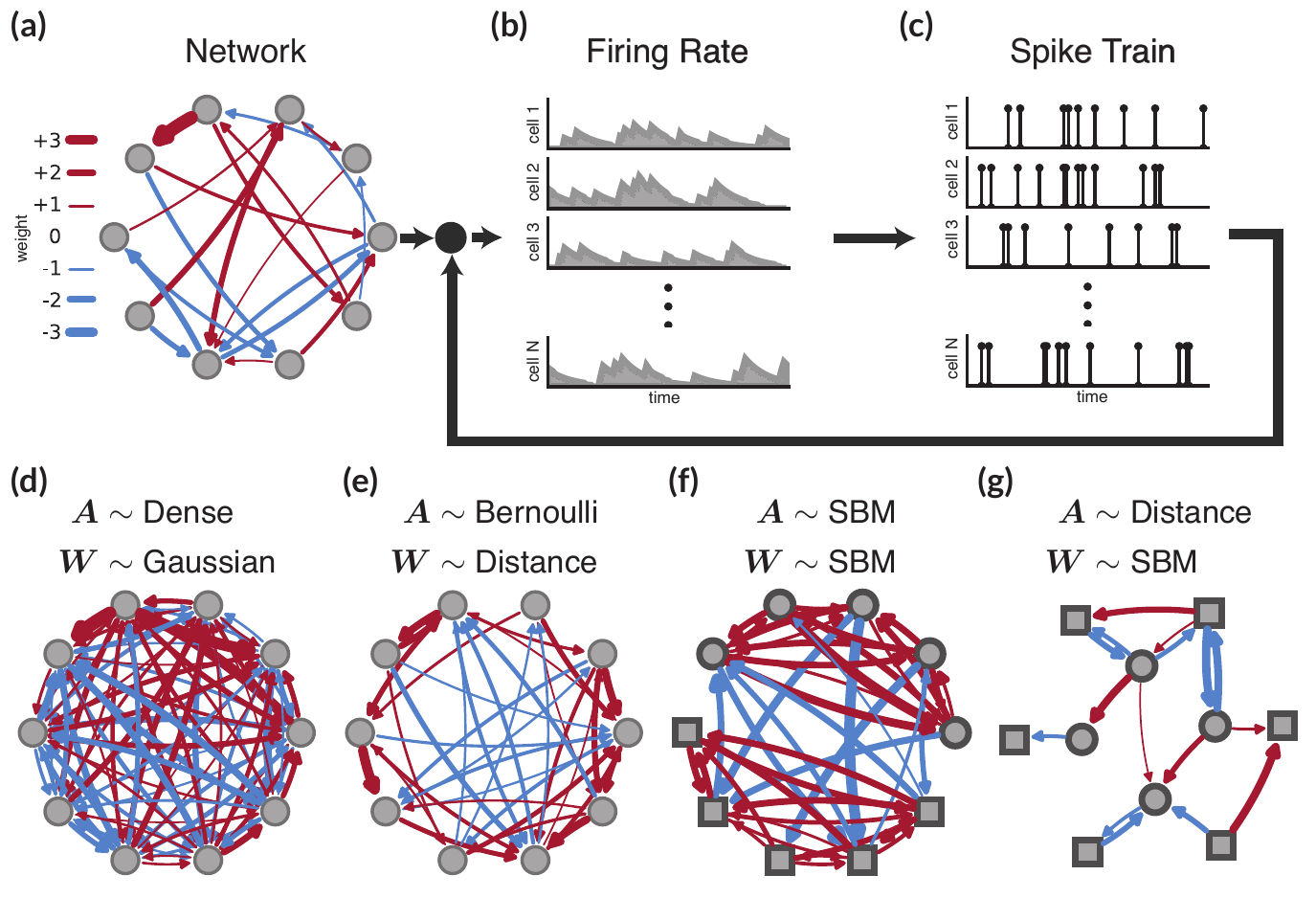}
  \end{subfigure}
  \caption[Components of the network GLM]{ Components of the
    generative model.  \textbf{(a)} Neurons influence one another via
    a sparse weighted network of interactions.  \textbf{(b)} The
    network parameterizes an autoregressive model with a time-varying
    activation.  \textbf{(c)} Spike counts are randomly drawn from a
    discrete distribution with a logistic link function. Each spike
    induces an impulse response on the activation of downstream
    neurons.  \textbf{(d)} Standard GLM analyses correspond to a
    fully-connected network with Gaussian or Laplace distributed
    weights, depending on the regularization.  \textbf{(e-g)} In this
    work, we consider structured models like the stochastic block
    model (SBM), in which neurons have discrete latent types
    (e.g. \emph{square} or \emph{circle}), and the latent distance
    model, in which neurons have latent locations that determine their
    probability of connection, capturing intuitive and interpretable
    patterns of connectivity.  }
  \label{fig:fig1}
  \vspace{-1em}
\end{figure}

\section{Probabilistic Model}
\label{sec:model}

Figure~\ref{fig:fig1} illustrates the components of our framework.  We
begin with a prior distribution on networks that generates a set of
weighted connections between neurons (Fig.~\ref{fig:fig1}a).  A directed edge
indicates a functional relationship between the spikes of
one neuron and the activation of its downstream neighbor. Each
spike induces a weighted impulse response on the activation of the
downstream neuron~(Fig.~\ref{fig:fig1}b).  The activation is
converted into a nonnegative firing rate from which spikes are
stochastically sampled (Fig.~\ref{fig:fig1}c). These spikes then feed
back into the subsequent activation, completing an autoregressive loop,
the hallmark of the GLM \citep{Paninski-2004,
  Truccolo-2005}. Models like these have provided valuable insight
into complex population recordings \citep{Pillow-2008}. We detail
the three components of this model in the reverse order, working backward from
the observed spike counts through the activation to the underlying network.

\subsection{Logistic Spike Count Models}
Generalized linear models assume a stochastic spike generation
mechanism.  Consider a matrix of spike
counts,~${\bS \in \naturals^{T \times N}}$, for~$T$
time bins and~$N$ neurons.  The expected number of spikes fired by the~$n$-th neuron
in the~$t$-th time bin,~$\bbE[s_{t,n}]$, is modeled as a nonlinear
function of the instantaneous \emph{activation},~$\psi_{t,n}$, and a
static, neuron-specific parameter,~$\nu_n$.
Table~\ref{tab:obs_models} enumerates the three spike count models
considered in this paper, all of which use the logistic
function,~${\sigma(\psi) =e^\psi (1+e^\psi)^{-1}}$, to rectify the
activation.  The Bernoulli distribution is appropriate for binary
spike counts, whereas the binomial and negative binomial have support
for~$s\in[0,\nu]$ and~$s \in [0, \infty)$, respectively.  Notably
lacking from this list is the Poisson distribution, which is not
directly amenable to the augmentation schemes we derive below;
however, both the binomial and negative binomial distributions
converge to the Poisson under certain limits.  Moreover, these
distributions afford the added flexibility of modeling under- and
over-dispersed spike counts, a biologically significant feature of
neural spiking data~\citep{goris2014partitioning}.  Specifically,
while the Poisson has unit dispersion (its mean is equal to its
variance), the binomial distribution is always under-dispersed, since
its mean always exceeds its variance, and the negative binomial is
always over-dispersed, with variance greater than its mean.

Importantly, all of these distributions can be written in a standard
form, as shown in Table~\ref{tab:obs_models}. We exploit this fact to
develop an efficient Markov chain Monte Carlo (MCMC) inference
algorithm described in Section~\ref{sec:inference}.

% Firing rate \lambda_{t,n} = g(\psi_{t,n})
\begin{table}
\begin{center}
\begin{tabular}{c|c|c|c|c}
  Distribution & $p(s \given \psi, \nu)$ & Standard Form & $\bbE[s]$ & $\Var(s)$ \\
  \hline
  $\distBernoulli(\sigma(\psi))$ & $\sigma(\psi)^s \, \sigma(-\psi)^{1-s}$
  & $\frac{(e^\psi)^s}{1+e^\psi}$
  & $\sigma(\psi)$ & $\sigma(\psi) \, \sigma(-\psi)$ \\
  $\distBinomial(\nu, \sigma(\psi))$ & ${\nu \choose s} \, \sigma(\psi)^s \, \sigma(-\psi)^{\nu-s}$
  & ${\nu \choose s} \,\frac{(e^\psi)^s}{(1+e^\psi)^\nu}$
  & $\nu \sigma(\psi)$ & $\nu \sigma(\psi) \, \sigma(-\psi)$ \\
  $\distNegBinomial(\nu, \sigma(\psi))$ & ${\nu + s -1 \choose s} \, \sigma(\psi)^s \, \sigma(-\psi)^{\nu}$
  & ${\nu +s - 1 \choose s} \,\frac{(e^\psi)^s}{(1+e^\psi)^{\nu+s}}$
  & $\nu e^\psi$ & $\nu e^\psi / \sigma(-\psi)$ \\
\end{tabular}
\end{center}
\caption{Table of conditional spike count distributions, their parameterizations,
  and their properties.}
\label{tab:obs_models}
\vspace{-1em}
\end{table}

% Activation Model
\subsection{Linear Activation Model}
The instantaneous activation of neuron~$n$ at time~$t$ is modeled as a
linear, autoregressive function of preceding spike counts of neighboring
neurons,
\begin{align}
  \psi_{t,n}
  &\triangleq b_n +  \sum_{m=1}^N \sum_{\Delta t = 1}^{\Delta t_{\mathsf{max}}} 
  h_{m \to n}[\Delta t] \cdot s_{t - \Delta t, m},
\end{align}
where~$b_n$ is the baseline activation of neuron~$n$ 
and~${h_{m \to n}: \{1, \ldots, \Delta t_{\mathsf{max}}\} \to \reals}$ is an impulse response
function that models the influence spikes on neuron~$m$ have on the activation of neuron
~$n$ at a delay of~$\Delta t$. To model the impulse response, we use a spike-and-slab
formulation \citep{Mitchell1988},
\begin{align}
  h_{m \to n}[\Delta t] &= a_{m \to n} \sum_{k=1}^K w_{m \to n}^{(k)} \, \phi_k[\Delta t].
\end{align}
Here, ~${a_{m \to n} \in \{0,1\}}$ is a binary variable indicating the
presence or absence of a connection from neuron~$m$ to neuron~$n$, the
weight~${\bw_{m \to n} =[w_{m \to n}^{(1)}, ..., w_{m \to n}^{(K)}]}$
denotes the strength of the connection, and~${\{\phi_k\}_{k=1}^K}$ is
a collection of fixed basis functions.  In this paper, we consider
scalar weights~${(K=1)}$ and use an exponential basis
function,~$\phi_1[\Delta t]=e^{-\Delta t / \tau}$, with time constant
of~${\tau=15}$ms.  Since the basis function and the spike train are
fixed, we precompute the convolution of the spike train and the basis
function to obtain~${\widehat{s}_{t,m}^{(k)} = \sum_{\Delta t =
    1}^{\Delta t_{\mathsf{max}}} \phi_k[\Delta t] \cdot s_{t-\Delta t,
    m}}$.  Finally, we combine the connections, weights, and filtered
spike trains and write the activation as,
\begin{align}
  \label{eq:linear_activation}
  \psi_{t,n} &= (\ba_{n} \odot \bw_n)^\trans \, \widehat{\bs}_t,
\end{align}
where~${\ba_{n} = [ 1, a_{1 \to n} \bone_K, ..., a_{N \to n} \bone_K ]}$, 
~${\bw_n =
  [b_n, \bw_{1 \to n}, ..., \bw_{N \to n}]}$, 
and~${\widehat{\bs}_t = [1,
    \widehat{s}_{t,1}^{(1)}, ..., \widehat{s}_{t,N}^{(K)}
]}$. Here,~$\odot$ denotes the Hadamard (elementwise)
product and~$\bone_K$ is
length-$K$ vector of ones. Hence, all of these vectors are of size~${1+NK}$.  The
difference between our formulation and the standard GLM is that we
have explicitly modeled the sparsity of the weights in~$a_{m \to n}$.
In typical formulations
\citep[e.g.,][]{Pillow-2008}, all connections are present
and the weights are regularized
with~$\ell_1$ and~$\ell_2$ penalties to promote sparsity. 
Instead, we consider structured approaches to modeling the sparsity
and weights.

% Network model
\subsection{Random Network Models}
Patterns of functional interaction can provide great insight into the 
computations performed by neural circuits. Indeed, many circuits are 
informally described in terms of ``types'' of neurons that 
perform a particular role, or the ``features'' that neurons encode. 
Random network models formalize these intuitive descriptions. Types and 
features correspond to latent variables in a probabilistic model that 
governs how likely neurons are to connect and how strongly they influence
each other. 

Let~${\bA = \{\{a_{m \to n}\}\}}$ and~${\bW = \{\{\bw_{m \to n}\}\}}$ denote 
the binary adjacency matrix and the real-valued array of weights, respectively.
Now suppose~$\{\bu_n\}_{n=1}^N$ and~$\{\bv_n\}_{n=1}^N$ are sets of
neuron-specific latent variables that govern the distributions over~$\bA$
and~$\bW$. Given these latent variables and global
parameters~$\btheta$, the entries in~$\bA$ are conditionally
independent Bernoulli random variables, and the entries in~$\bW$ are
conditionally independent Gaussians. That is,
\begin{multline}
  p(\bA, \bW \given \{\bu_n, \bv_n\}_{n=1}^N, \btheta)
  =
  \prod_{m=1}^N \prod_{n=1}^N
  \distBernoulli \left(a_{m \to n} \given \rho(\bu_m, \bu_n, \btheta) \right) \\
  \times
  \distNormal \left( \bw_{m \to n} \given \bmu(\bv_m, \bv_n, \btheta), \bSigma(\bv_m, \bv_n, \btheta) \right),
\end{multline}
where~$\rho(\cdot)$,~$\bmu(\cdot)$, and~$\bSigma(\cdot)$ are functions that
output a probability, a mean vector, and a covariance matrix, respectively.
We recover the standard GLM when~${\rho(\cdot) \equiv 1}$, but here we can 
take advantage of structured priors like the stochastic block model (SBM)
\citep{Nowicki-2001}, in which each neuron has a discrete type,
 and the latent distance model~\citep{Hoff2008}, in which each neuron has a 
latent location. Table~\ref{tab:network_models} outlines the various 
models considered in this paper.

We can mix and match these models as shown in
Figure~\ref{fig:fig1}(d-g). For example, in Fig.~\ref{fig:fig1}g, the
adjacency matrix is distance-dependent and the weights are block
structured.  Thus, we have a flexible language for expressing
hypotheses about patterns of interaction. In fact, the simple models
enumerated above are instances of a rich family of exchangeable
networks known as Aldous-Hoover random graphs, which have been
recently reviewed by~\citet{orbanz2015bayesian}.

\begin{table}
\begin{center}
\begin{tabular}{c||c||c|c}
Name & $\rho(\bu_m, \bu_n, \btheta)$ & $\bmu(\bv_m, \bv_n, \btheta)$ & $\bSigma(\bv_m, \bv_n, \btheta)$\\
\hline
Dense Model & $1$ & $\overline{\bmu}$ & $\overline{\bSigma}$ \\
Independent Model & $\overline{\rho}$ & $\overline{\bmu}$ & $\overline{\bSigma}$ \\
Stochastic Block Model & $\rho_{\bu_m \to \bu_n}$ & $\bmu_{\bv_m \to \bv_n}$ & $\bSigma_{\bv_m \to \bv_n}$ \\
Latent Distance Model & $\sigma(-||\bu_n - \bv_m||_2^2 + \gamma_0)$ & $-||\bv_n - \bv_m||_2^2 + \mu_0$ & $\eta^2$ \\
\hline 
\end{tabular}
\end{center}
\caption{Random network models for the binary adjacency matrix or the Gaussian weight matrix.}
\label{tab:network_models}
\vspace{-1em}
\end{table}

\section{Bayesian Inference}
\label{sec:inference}

Generalized linear models are often fit via maximum \emph{a
posteriori} (MAP) estimation \citep{Paninski-2004, Truccolo-2005,
  Pillow-2008, vidne2012modeling}. However, as we scale to larger populations of neurons,
there will inevitably be structure in the posterior that is not
reflected with a point estimate.  Technological advances are expanding the number of neurons that can be recorded simultaneously, but ``high-throughput'' recording of many individuals is still a distant hope.  Therefore we expect the complexities of our models to expand faster than the available distinct data sets to fit them.  In this situation, accurately capturing uncertainty is critical.  Moreover, in the Bayesian framework, we also have a coherent way to perform model selection and evaluate hypotheses regarding complex underlying structure.  Finally, after introducing a binary
adjacency matrix and hierarchical network priors, the log posterior
is no longer a concave function of model parameters, making direct
optimization challenging (though see~\citet{soudry2013shotgun} for
recent advances in tackling similar problems). These considerations motivate a fully Bayesian approach.

Computation in rich Bayesian models is often challenging, but through thoughtful modeling decisions it is sometimes possible to find representations that lead to efficient inference.  In this case, we have carefully chosen the logistic models of the preceding section in order to make it possible to apply the \polyagamma augmentation scheme \citep{polson2013bayesian}.
The principal advantage of this approach is that,
given the\polyagamma auxiliary variables, the conditional distribution
of the weights is Gaussian, and hence is amenable to efficient Gibbs
sampling. Recently, \citet{Pillow2012} used this technique to develop
inference algorithms for negative binomial factor analysis models of
neural spike trains.  We build on this work and show how this
conditionally Gaussian structure can be exploited to derive efficient,
collapsed Gibbs updates.

\subsection{Collapsed Gibbs updates for Gaussian observations}
Suppose the observations were actually Gaussian distributed, i.e. 
~${s_{t,n} \sim \distNormal(\psi_{t,n}, \nu_n)}$. 
The most
challenging aspect of inference is then sampling the posterior distribution
over discrete connections,~$\bA$. There may be many posterior modes
corresponding to different patterns of connectivity. Moreover,~$a_{m \to n}$
and~$\bw_{m \to n}$ are often highly correlated, which leads to poor 
mixing of na\"ive Gibbs sampling.
Fortunately, when the
observations are Gaussian, we may integrate over possible weights and
sample the binary adjacency matrix from its collapsed conditional
distribution.

We combine the conditionally independent Gaussian priors
on~$\{\bw_{m \to n}\}$ and~$b_n$ into a joint Gaussian
distribution,~${\bw_n \given \{\bv_n\}, \btheta \sim \distNormal(\bw_n
  \given \bmu_n, \bSigma_n)}$,
where~$\bSigma_n$ is a
block diagonal covariance matrix.  Since~$\psi_{t,n}$ is linear
in~$\bw_n$ (see Eq.~\ref{eq:linear_activation}), a Gaussian likelihood
is conjugate with this Gaussian prior, given~$\ba_n$
and~$\widehat{\bS}=\{\widehat{\bs}_t\}_{t=1}^T$. This yields the
following closed-form conditional:
\begin{align*}
%  \label{eq:w_conditional}
  p(\bw_n \given \widehat{\bS}, \ba_n, \bmu_n, \bSigma_n)
  \propto
  \distNormal(\bw_n \given \bmu_n, \bSigma_n) \,
  \prod_{t=1}^T \distNormal(s_{t,n} \given (\ba_n \odot \bw_n)^\trans \, \widehat{\bs}_t, \, \nu_n) 
 \propto \distNormal(\bw_n \given \widetilde{\bmu}_n, \widetilde{\bSigma}_n), \\
  \widetilde{\bSigma}_n = \left[ \bSigma_n^{-1} +
  \left(\widehat{\bS}^\trans (\nu_n^{-1} \bI) \widehat{\bS} \right) \odot (\ba_n \ba_n^\trans) \right]^{-1}, 
  \quad
  \widetilde{\bmu}_n = \widetilde{\bSigma}_n \left[ \bSigma_n^{-1} \bmu_n +
  \left(\widehat{\bS}^\trans (\nu_n^{-1} \bI)\bs_{:,n} \right) \odot \ba_n \right].
\end{align*}

Now, consider the conditional distribution of~$\ba_n$, integrating out
the corresponding weights. The prior distribution over~$\ba_n$ is a
product of Bernoulli distributions with
parameters~${\brho_n = \{\rho(\bu_m, \bu_n, \btheta)\}_{m=1}^N}$.
The conditional distribution is proportional to the
ratio of the prior and posterior partition functions,
\begin{align*}
  \nonumber
  p(\ba_n \given \widehat{\bS}, \brho_n, \bmu_n, \bSigma_n)
  &= \int p(\ba_n, \bw_n \given \widehat{\bS}, \brho_n, \bmu_n, \bSigma_n) \, \mathrm{d} \bw_n  \\
%  \label{eq:a_conditional}
  &= p(\ba_n \given \brho_n) \, \frac{\big| \bSigma_n \big|^{-\frac{1}{2}} \exp \Big \{-\frac{1}{2} \bmu_n^\trans \bSigma_n^{-1} \bmu_n \Big \} }
  {\big| \widetilde{\bSigma}_n \big|^{-\frac{1}{2}} \exp \Big \{-\frac{1}{2} \widetilde{\bmu}_n^\trans \widetilde{\bSigma}_n^{-1} \widetilde{\bmu}_n \Big \}}.
\end{align*}
Thus, we perform a joint update of~$\ba_n$ and~$\bw_n$ by
collapsing out the weights to directly sample the binary
entries of~$\ba_n$. We iterate over each entry,~$a_{m \to n}$,
and sample it from its conditional distribution given~$\{a_{m' \to n}\}_{m' \neq m}$.
Having sampled~$\ba_n$, we sample~$\bw_n$ from its Gaussian conditional.

%Thus, we can efficiently sample from the conditional
%distribution of~$\ba_n$ and~$\bw_n$ by first iterating
%over each neuron~${m \in \{1, \ldots, N\}}$ and sampling
%a new value of~$a_{n \from m}$, fixing the values of~$a_{n \from m'}$
%for~$m' \neq m$ and integrating out the value of~$\bw_n$.
%To do so, we simply evaluate the marginal probability in Eq.~\ref{eq:a_conditional}
%for both values of~$a_{n \from m}$ and resample accordingly.
%Moreover, note that if~$a_{n \from m}$, the log determinant and the quadratic
%form in the numerator of Eq.~\ref{eq:a_conditional} will cancel with the
%corresponding term in the denominator. Thus, if~$\bA$ is $d$-sparse (i.e. 
%each neuron  has at most~$d$ incoming edges) evaluating
%the marginal probability of~$\ba_n$ has complexity an~$O(d^3)$.
%Once~$\ba_n$ has been completely
%resampled, we can sample a new value of~$\bw_n$ from its multivariate
%Gaussian conditional distribution, given by~Eq.~\ref{eq:w_conditional}.

\subsection{\polyagamma augmentation for discrete observations}
Now, let us turn to the non-conjugate case of discrete count
observations.  The \polyagamma augmentation
\citep{polson2013bayesian} introduces auxiliary
variables,~$\omega_{t,n}$, conditioned upon which the discrete
likelihood appears Gaussian and our collapsed Gibbs updates apply.
The integral identity underlying this scheme is
\begin{align}
\label{eq:pg_identity}
c \, \frac{(e^{\psi})^a}{(1+e^{\psi})^b} = c \, 2^{-b} e^{\kappa \psi}
\int_{0}^{\infty} e^{-\omega \psi^2 /2} \, p_{\mathrm{PG}}(\omega
\given b, 0) \, \mathrm{d}\omega,
\end{align}
where~${\kappa=a-b/2}$ and~$p(\omega\given b, 0)$ is the density of
the P\'{o}lya-gamma distribution~${\distPolyaGamma(b, 0)}$, which does
not depend on $\psi$.  Notice that the discrete likelihoods in
Table~\ref{tab:obs_models} can all be rewritten like the left-hand
side of~\eqref{eq:pg_identity}, for some~$a$,~$b$, and~$c$ that are
functions of~$s$ and~$\nu$.  Using~\eqref{eq:pg_identity} along with
priors~$p(\psi)$ and~$p(\nu)$, we write the joint density of
$(\psi, s, \nu)$ as
\begin{align}
  \label{eq:pg_joint}
  p(s, \nu, \psi)
%  &= p(\nu) \, p(\psi) \, c(s, \nu) \frac{(e^\psi)^{a(s, \nu)}}{(1+e^\psi)^{b(s, \nu)}} \\
%  \nonumber
  &= \int_0^\infty
  p(\nu) \, p(\psi) \, c(s, \nu) \, 2^{-b(s, \nu)} e^{\kappa(s, \nu) \psi} e^{-\omega \psi^2/2} \, p_{\mathrm{PG}}(\omega \given b(s, \nu), 0) \; \mathrm{d}\omega.
\end{align}
The integrand of Eq.~\ref{eq:pg_joint} defines a joint density on $(s,
\nu, \psi, \omega)$ which admits $p(s, \nu, \psi)$ as a marginal
density.  Conditioned on the auxiliary variable, $\omega$, the
likelihood as a function of~$\psi$ is,
\begin{align*}
  p(s \given \psi, \nu, \omega)
  &\propto e^{\kappa(s, \nu) \psi} e^{- \omega \psi^2/2} 
\propto \distNormal \left( \omega^{-1}  \kappa(s, \nu) \given \psi, \,  \omega^{-1} \right).
\end{align*}
Thus, after conditioning on~$s$,~$\nu$, and~$\omega$, we effectively have a
linear Gaussian likelihood for~$\psi$.

We apply this augmentation
scheme to the full model, introducing auxiliary
variables,~$\omega_{t,n}$ for each spike count,~$s_{t,n}$.  Given
these variables, the conditional distribution of~$\bw_n$ can be
computed in closed form, as before.  Let~${\bkappa_n =
  [\kappa(s_{1,n}, \nu_n), \ldots, \kappa(s_{T,n}, \nu_n)]}$ and
~${\bOmega_n = \diag([\omega_{1,n}, \ldots, \omega_{T,n}] )}$.  Then
we have~$ p(\bw_n \given \bs_n, \widehat{\bS}, \ba_n, \bmu_n,
\bSigma_n, \bomega_n, \nu_n) \propto \distNormal(\bw_n \given
\widetilde{\bmu}_n, \widetilde{\bSigma}_n)$, where
\begin{align*}
  \widetilde{\bSigma}_n = \left[ \bSigma_n^{-1} +
  \left(\widehat{\bS}^\trans \bOmega_n \widehat{\bS} \right) \odot (\ba_n \ba_n^\trans) \right]^{-1}, \quad
  \widetilde{\bmu}_n = \widetilde{\bSigma}_n \left[ \bSigma_n^{-1} \bmu_n +
  \left(\widehat{\bS}^\trans \bkappa_n \right) \odot \ba_n \right].
\end{align*}

%% Synth connection recovery example
\begin{figure}[t!]
  \centering
  \begin{subfigure}[b]{4.5in}
    \centering
    \includegraphics[width=\textwidth]{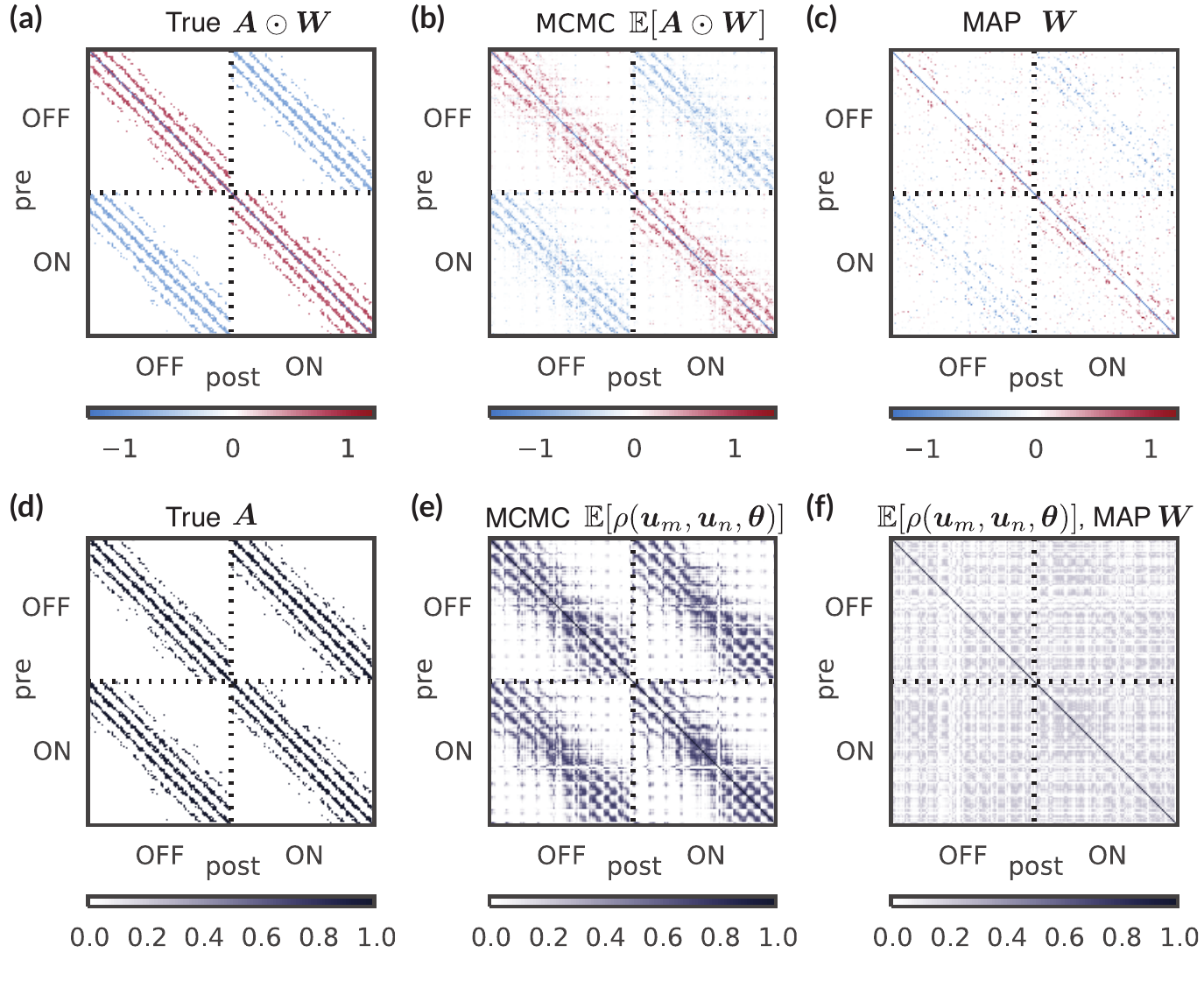}
  \end{subfigure}
  \vspace{-1em}
  \caption{ Weighted adjacency matrices showing inferred networks and connection probabilities for
    synthetic data.  \textbf{(a,d)} True network. \textbf{(b,e)} Posterior mean using joint inference of network GLM.
    \textbf{(c,f)} MAP estimation.  }
  \label{fig:synth}
  \vspace{-1em}
\end{figure}

Having introduced auxiliary variables, we must now derive
Markov transitions to update them as well. Fortunately, the
\polyagamma distribution is designed such that the conditional
distribution of the auxiliary variables is simply a ``tilted'' \polyagamma
distribution,
\begin{align*}
  p(\omega_{t,n} \given s_{t,n}, \nu_n, \psi_{t,n})
  &= p_{\mathrm{PG}}(\omega_{t,n} \given b(s_{t,n}, \nu_n), \, \psi_{t,n}).
\end{align*}
These auxiliary variables are conditionally independent given the
activation and hence can be sampled in parallel. Moreover, efficient
algorithms are available to generate \polyagamma random
variates~\citep{windle2014sampling}.
Our Gibbs updates for the remaining parameters and latent variables ($\nu_n$,~$\bu_n$,~$\bv_n$, and~$\btheta$)
are described in the supplementary material. 
A Python implementation
of our inference algorithm is available at~\url{https://github.com/slinderman/pyglm}.

\section{Synthetic Data Experiments}
\label{sec:synthetic}

The need for network models is most pressing in recordings of large
populations where the network is difficult to estimate and even harder
to interpret. To assess the robustness and scalability of our
framework, we apply our methods to simulated data with known ground
truth.  We simulate a one minute recording (1ms time bins) from a population of 200
neurons with discrete latent types that govern the connection strength
via a stochastic block model and continuous latent locations that
govern connection probability via a latent distance model. The spikes
are generated from a Bernoulli observation model.

First, we show that our approach of jointly inferring the network
and its latent variables can provide dramatic improvements over
alternative approaches. For comparison, consider the two-step
procedure of \citet{stevenson2009bayesian} in which the network is fit
with an $\ell_1$-regularized GLM and \textit{then} a probabilistic
network model is fit to the GLM connection weights. The advantage of
this strategy is that the expensive GLM fitting is only performed
once.  However, when the data is limited, both the network and the
latent variables are uncertain.  Our Bayesian approach finds a very
accurate network (Fig.~\ref{fig:synth}b) by jointly sampling
networks and latent variables. In contrast, the standard GLM does not
account for latent structure and finds strong connections as well as
spuriously correlated neurons~(Fig.~\ref{fig:synth}c).  Moreover, our fully
Bayesian approach finds a set of latent locations that mimics the 
true locations and therefore accurately estimates connection probability
(Fig.~\ref{fig:synth}e). In contrast, subsequently fitting a latent distance
model to the adjacency matrix of a thresholded GLM network finds an
embedding that has no resemblance to the true locations, which is reflected
in its poor estimate of connection probability (Fig.~\ref{fig:synth}f).

\begin{figure}[t!]
  \centering
  \begin{subfigure}[b]{4.5in}
    \centering
    \includegraphics[width=\textwidth]{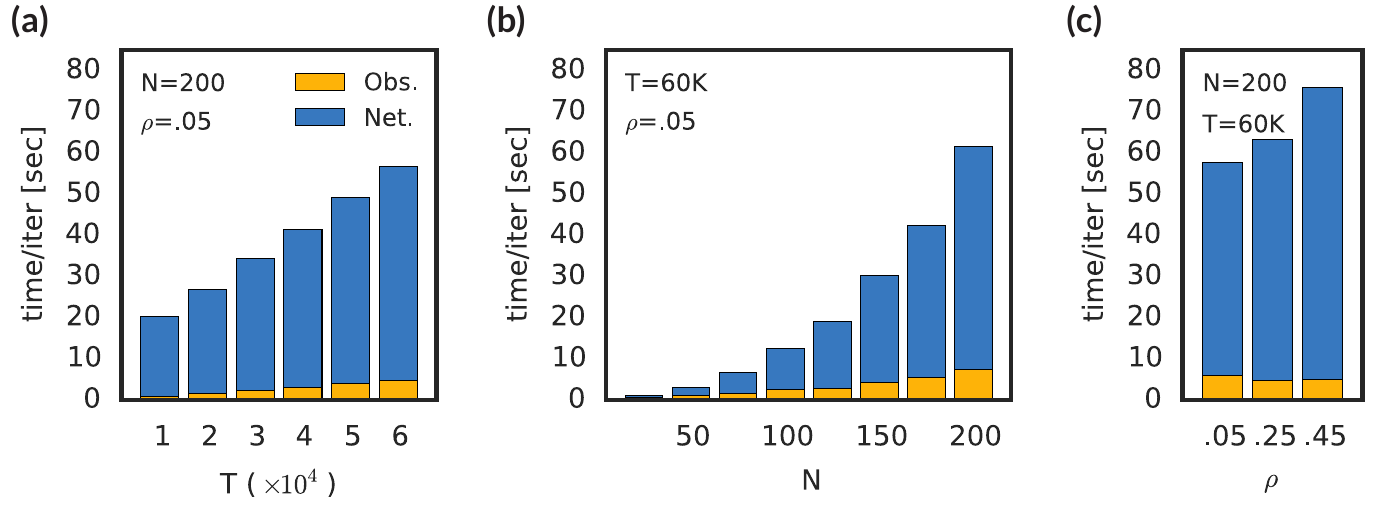}
  \end{subfigure}
  \vspace{-1em}
  \caption[Scalability of the proposed Bayesian inference algorithm]{
    Scalability of our inference algorithm as a function of: 
    \textbf{(a)} the number of time bins,~$T$;
    \textbf{(b)} the number of neurons,~$N$; and
    \textbf{(c)} the average sparsity of the network,~$\rho$.
    Wall-clock time is divided into time spent sampling auxiliary
    variables (``Obs.'') and time spent sampling the network (``Net.'').
  }
  \label{fig:scalability}
  \vspace{-1em}
\end{figure}

Next, we address the scalability of our MCMC algorithm.  
Three major parameters govern the complexity of inference: the
number of time bins,~$T$; the number of neurons,~$N$; and the level of
sparsity,~$\rho$. The following experiments were run on a quad-core
Intel i5 with 6GB of RAM.  As shown in Fig.~\ref{fig:scalability}a,
the wall clock time per iteration scales linearly with~$T$ since we
must resample~${NT}$ auxiliary variables. We scale at least quadratically
with~$N$ due to the network, as shown in~Fig.~\ref{fig:scalability}b.
However, the total cost could actually be worse than quadratic since
the cost of updating each connection could depend on~$N$. Fortunately,
the complexity of our collapsed Gibbs sampling algorithm only depends
on the number of incident connections,~$d$, or
equivalently, the sparsity~$\rho=d/N$.  Specifically, we must solve a 
linear system of size~$d$, which incurs a cubic cost, as seen in Fig.~\ref{fig:scalability}c.
%Finally, note that auxiliary variables
%can be resampled in parallel, as can the columns of~$\bA$ and~$\bW$. 
%Together, these allow for an~$O(T)$ and~$O(N)$ speedup, respectively.

\section{Retinal Ganglion Cells}
\label{sec:rgc}

Finally, we demonstrate the efficacy of this approach with an
application to spike trains simultaneously recorded from a population
of 27 retinal ganglion cells (RGCs), which have previously been
studied by~\citet{Pillow-2008}.  Retinal ganglion cells respond to
light shown upon their receptive field.  Thus, it is natural to
characterize these cells by the location of their receptive field
center.  Moreover, retinal ganglion cells come in a variety of
types~\citep{sanes2015types}. This population is comprised of two
types of cells, \textit{on} and \textit{off} cells, which are
characterized by their response to visual stimuli. \textit{On} cells
increase their firing when light is shone upon their receptive field;
\textit{off} cells decrease their firing rate in response to light in
their receptive field.  In this case, the population is driven by a
binary white noise stimulus. Given the stimulus, the cell locations
and types are readily inferred. Here, we show how these intuitive
representations can be discovered in a purely unsupervised manner
given one minute of spiking data alone and no knowledge of the
stimulus.

\begin{figure}[t!]
  \centering
  \begin{subfigure}[b]{5.5in}
    \centering
    \includegraphics[width=.95\textwidth]{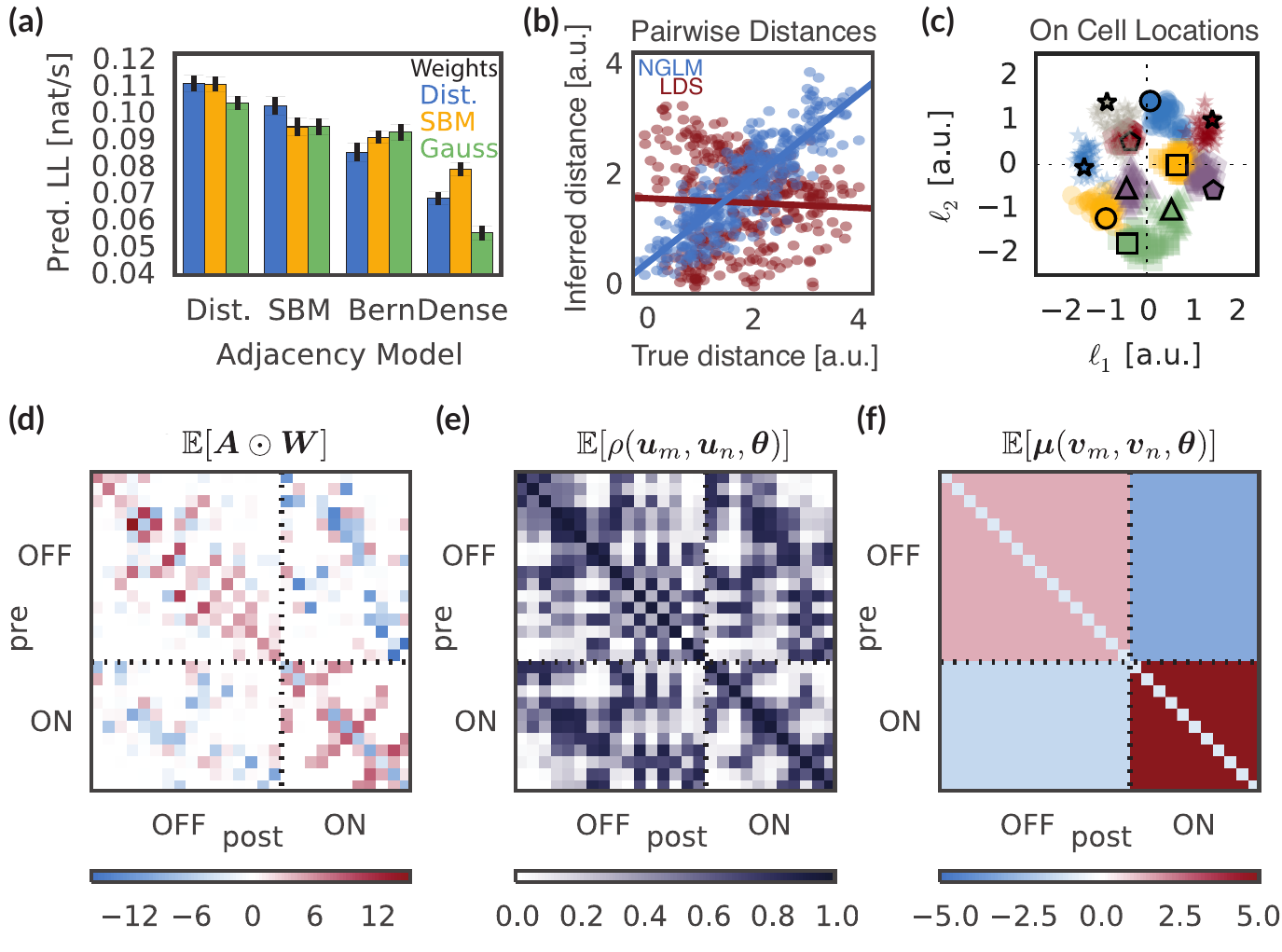}
  \end{subfigure}
  \vspace{-1em}
  \caption{Using our framework, retinal ganglion cell types and
    locations can be inferred from spike trains alone. \textbf{(a)} Model comparison.
    \textbf{(b)} True and inferred distances between cells. \textbf{(c)} True and inferred
    cell locations. \textbf{(d-f)} Inferred network, connection probability, and mean weight,
    respectively. See main text for further details.}
  \label{fig:rgc}
  \vspace{-1em}
\end{figure}

Figure~\ref{fig:rgc} illustrates the results of our analysis. Since
the data are binned at 1ms resolution, we have at most one spike per
bin and we use a Bernoulli observation model.  We fit the 12
network models of Table~\ref{tab:network_models} (4 adjacency models
and 3 weight models), 
%running our MCMC algorithm for 1000 iterations
%and using the last 500 to estimate posterior expectations. 
and we find that, 
in terms of predictive log
likelihood of held-out neurons, a latent distance model of
the adjacency matrix and SBM of the weight matrix performs best
(Fig.~\ref{fig:rgc}a).  See the supplementary material for a detailed
description of this comparison.  Looking into the latent locations
underlying the adjacency matrix our network GLM (NGLM), we find that
the inferred distances between cells are highly correlated with the
distances between the true locations. For comparison, we also fit a 2D
Bernoulli linear dynamical system (LDS) --- the Bernoulli equivalent
of the Poisson LDS \citep{macke2011empirical} --- and we take rows of
the $N{ \times 2}$~emission matrix as locations. In contrast to our
network GLM, the distances between LDS locations are nearly
uncorrelated with the true distances (Fig.~\ref{fig:rgc}b) since the
LDS does not capture the fact that distance only affects the
probability of connection, not the weight. Not only are our distances
accurate, the inferred locations are nearly identical to the true
locations, up to affine transformation. In Fig.~\ref{fig:rgc}c,
semitransparent markers show the inferred \textit{on} cell locations,
which have been rotated and scaled to best align with the true
locations shown by the outlined marks.  Based solely on patterns of
correlated spiking, we have recovered the receptive field arrangements.
%that characterize this population.

Fig.~\ref{fig:rgc}d shows the inferred network,~${\bA \odot \bW}$,
under a latent distance model of connection probability and a
stochastic block model for connection weight. 
%Neurons are sorted by type and then by $x$-location. 
The underlying connection probabilities from the distance model are shown
in Fig.~\ref{fig:rgc}e.  Finally, Fig.~\ref{fig:rgc}f shows that
we have discovered not only the cell locations, but also their latent
types. With an SBM, the mean weight is a function of latent type, and under
the posterior, the neurons are clearly clustered into the two true types that
exhibit the expected within-class excitation and between-class inhibition.

\vspace{-1em}
\section{Conclusion}
\label{sec:discussion}
%\vspace{-1em}
Our results with both synthetic and real neural data provide 
compelling evidence that our methods can find meaningful structure
underlying neural spike trains. Given the extensive work on characterizing retinal 
ganglion cell responses, we have considerable evidence that the 
representation we learn from spike trains alone is indeed the 
optimal way to summarize this population of cells. This 
lends us confidence that we may trust the representations learned from 
spike trains recorded from more enigmatic brain areas as well.
While we have omitted stimulus from our models and only used it for 
confirming types and locations, in practice we could incorporate it into
our model and even capture type- and location-dependent patterns of stimulus dependence with our hierarchical approach.
%Likewise, dynamic latent states plug directly into the linear activation.
Likewise, the network GLM could be combined with the PLDS as in \citet{vidne2012modeling}
to capture sources of low dimensional, shared variability.

Latent functional networks underlying spike trains can provide unique
insight into the structure of neural populations.  Looking forward,
methods that extract interpretable representations from complex neural
data, like those developed here, will be key to capitalizing on the
dramatic advances in neural recording technology. We have shown that
networks provide a natural bridge to connect neural types and features
to spike trains, and demonstrated promising results on both real and
synthetic data.

\begin{small}{
\textbf{Acknowledgments.}
We thank E. J. Chichilnisky, A. M. Litke, A. Sher and J. Shlens for retinal data.
SWL is supported by the Simons Foundation SCGB-418011.
RPA is supported by NSF IIS-1421780 and the Alfred P. Sloan Foundation.
JWP was supported by grants from the McKnight Foundation, Simons Collaboration on the Global Brain (SCGB AWD1004351), NSF CAREER Award (IIS-1150186), and NIMH grant MH099611.
}
\end{small}

\bibliographystyle{abbrvnat}
\small
\bibliography{arxiv}

\begin{thebibliography}{23}
\providecommand{\natexlab}[1]{#1}
\providecommand{\url}[1]{\texttt{#1}}
\expandafter\ifx\csname urlstyle\endcsname\relax
  \providecommand{\doi}[1]{doi: #1}\else
  \providecommand{\doi}{doi: \begingroup \urlstyle{rm}\Url}\fi

\bibitem[Ahrens et~al.(2013)Ahrens, Orger, Robson, Li, and
  Keller]{ahrens2013whole}
M.~B. Ahrens, M.~B. Orger, D.~N. Robson, J.~M. Li, and P.~J. Keller.
\newblock Whole-brain functional imaging at cellular resolution using
  light-sheet microscopy.
\newblock \emph{Nature methods}, 10\penalty0 (5):\penalty0 413--420, 2013.

\bibitem[Brillinger et~al.(1976)Brillinger, Bryant~Jr, and
  Segundo]{brillinger1976identification}
D.~R. Brillinger, H.~L. Bryant~Jr, and J.~P. Segundo.
\newblock Identification of synaptic interactions.
\newblock \emph{Biological Cybernetics}, 22\penalty0 (4):\penalty0 213--228,
  1976.

\bibitem[Gerhard et~al.(2013)Gerhard, Kispersky, Gutierrez, Marder, Kramer, and
  Eden]{Gerhard-2013}
F.~Gerhard, T.~Kispersky, G.~J. Gutierrez, E.~Marder, M.~Kramer, and U.~Eden.
\newblock Successful reconstruction of a physiological circuit with known
  connectivity from spiking activity alone.
\newblock \emph{PLoS Computational Biology}, 9\penalty0 (7):\penalty0 e1003138,
  2013.

\bibitem[Goris et~al.(2014)Goris, Movshon, and
  Simoncelli]{goris2014partitioning}
R.~L. Goris, J.~A. Movshon, and E.~P. Simoncelli.
\newblock Partitioning neuronal variability.
\newblock \emph{Nature Neuroscience}, 17\penalty0 (6):\penalty0 858--865, 2014.

\bibitem[Grewe et~al.(2010)Grewe, Langer, Kasper, Kampa, and
  Helmchen]{grewe2010high}
B.~F. Grewe, D.~Langer, H.~Kasper, B.~M. Kampa, and F.~Helmchen.
\newblock High-speed in vivo calcium imaging reveals neuronal network activity
  with near-millisecond precision.
\newblock \emph{Nature methods}, 7\penalty0 (5):\penalty0 399--405, 2010.

\bibitem[Hoff(2008)]{Hoff2008}
P.~D. Hoff.
\newblock Modeling homophily and stochastic equivalence in symmetric relational
  data.
\newblock \emph{Advances in Neural Information Processing Systems 20},
  20:\penalty0 1--8, 2008.

\bibitem[Macke et~al.(2011)Macke, Buesing, Cunningham, Byron, Shenoy, and
  Sahani]{macke2011empirical}
J.~H. Macke, L.~Buesing, J.~P. Cunningham, M.~Y. Byron, K.~V. Shenoy, and
  M.~Sahani.
\newblock Empirical models of spiking in neural populations.
\newblock In \emph{Advances in neural information processing systems}, pages
  1350--1358, 2011.

\bibitem[Mitchell and Beauchamp(1988)]{Mitchell1988}
T.~J. Mitchell and J.~J. Beauchamp.
\newblock {Bayesian Variable Selection in Linear Regression}.
\newblock \emph{Journal of the American Statistical Association}, 83\penalty0
  (404):\penalty0 1023----1032, 1988.

\bibitem[Neal(2010)]{Neal10}
R.~M. Neal.
\newblock {MCMC} using {Hamiltonian} dynamics.
\newblock \emph{Handbook of Markov Chain Monte Carlo}, pages 113--162, 2010.

\bibitem[Nowicki and Snijders(2001)]{Nowicki-2001}
K.~Nowicki and T.~A.~B. Snijders.
\newblock Estimation and prediction for stochastic blockstructures.
\newblock \emph{Journal of the American Statistical Association}, 96\penalty0
  (455):\penalty0 1077--1087, 2001.

\bibitem[Orbanz and Roy(2015)]{orbanz2015bayesian}
P.~Orbanz and D.~M. Roy.
\newblock Bayesian models of graphs, arrays and other exchangeable random
  structures.
\newblock \emph{Pattern Analysis and Machine Intelligence, IEEE Transactions
  on}, 37\penalty0 (2):\penalty0 437--461, 2015.

\bibitem[Paninski(2004)]{Paninski-2004}
L.~Paninski.
\newblock {Maximum likelihood estimation of cascade point-process neural
  encoding models}.
\newblock \emph{Network: Computation in Neural Systems}, 15\penalty0
  (4):\penalty0 243--262, Jan. 2004.

\bibitem[Pillow and Scott(2012)]{Pillow2012}
J.~W. Pillow and J.~Scott.
\newblock Fully bayesian inference for neural models with negative-binomial
  spiking.
\newblock In F.~Pereira, C.~Burges, L.~Bottou, and K.~Weinberger, editors,
  \emph{Advances in Neural Information Processing Systems 25}, pages
  1898--1906. 2012.

\bibitem[Pillow et~al.(2008)Pillow, Shlens, Paninski, Sher, Litke,
  Chichilnisky, and Simoncelli]{Pillow-2008}
J.~W. Pillow, J.~Shlens, L.~Paninski, A.~Sher, A.~M. Litke, E.~Chichilnisky,
  and E.~P. Simoncelli.
\newblock Spatio-temporal correlations and visual signalling in a complete
  neuronal population.
\newblock \emph{Nature}, 454\penalty0 (7207):\penalty0 995--999, 2008.

\bibitem[Polson et~al.(2013)Polson, Scott, and Windle]{polson2013bayesian}
N.~G. Polson, J.~G. Scott, and J.~Windle.
\newblock Bayesian inference for logistic models using {P}{\'o}lya--gamma
  latent variables.
\newblock \emph{Journal of the American Statistical Association}, 108\penalty0
  (504):\penalty0 1339--1349, 2013.

\bibitem[Prevedel et~al.(2014)Prevedel, Yoon, Hoffmann, Pak, Wetzstein, Kato,
  Schr{\"o}del, Raskar, Zimmer, Boyden, et~al.]{prevedel2014simultaneous}
R.~Prevedel, Y.-G. Yoon, M.~Hoffmann, N.~Pak, G.~Wetzstein, S.~Kato,
  T.~Schr{\"o}del, R.~Raskar, M.~Zimmer, E.~S. Boyden, et~al.
\newblock Simultaneous whole-animal 3d imaging of neuronal activity using
  light-field microscopy.
\newblock \emph{Nature methods}, 11\penalty0 (7):\penalty0 727--730, 2014.

\bibitem[Sanes and Masland(2015)]{sanes2015types}
J.~R. Sanes and R.~H. Masland.
\newblock The types of retinal ganglion cells: current status and implications
  for neuronal classification.
\newblock \emph{Annual review of neuroscience}, 38:\penalty0 221--246, 2015.

\bibitem[Soudry et~al.(2013)Soudry, Keshri, Stinson, Oh, Iyengar, and
  Paninski]{soudry2013shotgun}
D.~Soudry, S.~Keshri, P.~Stinson, M.-h. Oh, G.~Iyengar, and L.~Paninski.
\newblock A shotgun sampling solution for the common input problem in neural
  connectivity inference.
\newblock \emph{arXiv preprint arXiv:1309.3724}, 2013.

\bibitem[Stevenson et~al.(2009)Stevenson, Rebesco, Hatsopoulos, Haga, Miller,
  and K{\"o}rding]{stevenson2009bayesian}
I.~H. Stevenson, J.~M. Rebesco, N.~G. Hatsopoulos, Z.~Haga, L.~E. Miller, and
  K.~P. K{\"o}rding.
\newblock Bayesian inference of functional connectivity and network structure
  from spikes.
\newblock \emph{Neural Systems and Rehabilitation Engineering, IEEE
  Transactions on}, 17\penalty0 (3):\penalty0 203--213, 2009.

\bibitem[Truccolo et~al.(2005)Truccolo, Eden, Fellows, Donoghue, and
  Brown]{Truccolo-2005}
W.~Truccolo, U.~T. Eden, M.~R. Fellows, J.~P. Donoghue, and E.~N. Brown.
\newblock A point process framework for relating neural spiking activity to
  spiking history, neural ensemble, and extrinsic covariate effects.
\newblock \emph{Journal of Neurophysiology}, 93\penalty0 (2):\penalty0
  1074--1089, 2005.
\newblock \doi{10.1152/jn.00697.2004}.

\bibitem[Vidne et~al.(2012)Vidne, Ahmadian, Shlens, Pillow, Kulkarni, Litke,
  Chichilnisky, Simoncelli, and Paninski]{vidne2012modeling}
M.~Vidne, Y.~Ahmadian, J.~Shlens, J.~W. Pillow, J.~Kulkarni, A.~M. Litke,
  E.~Chichilnisky, E.~Simoncelli, and L.~Paninski.
\newblock Modeling the impact of common noise inputs on the network activity of
  retinal ganglion cells.
\newblock \emph{Journal of computational neuroscience}, 33\penalty0
  (1):\penalty0 97--121, 2012.

\bibitem[Windle et~al.(2014)Windle, Polson, and Scott]{windle2014sampling}
J.~Windle, N.~G. Polson, and J.~G. Scott.
\newblock Sampling {P}{\'o}lya-gamma random variates: alternate and approximate
  techniques.
\newblock \emph{arXiv preprint arXiv:1405.0506}, 2014.

\bibitem[Zhou et~al.(2012)Zhou, Li, Carin, and Dunson]{Zhou2012}
M.~Zhou, L.~Li, L.~Carin, and D.~B. Dunson.
\newblock Lognormal and gamma mixed negative binomial regression.
\newblock In \emph{Proceedings of the 29th International Conference on Machine
  Learning}, pages 1343--1350, 2012.

\end{thebibliography}

\appendix

\section{Model comparison via Predictive Log Likelihood}
How can we compare different network models in a principled manner?
Predictive log likelihood of held-out time bins is insufficient since
it only depends directly on~$\bA$ and~$\bW$.  The network prior does
aid in the estimation of~$\bA$ and~$\bW$, but this is only an indirect
effect, and it may be small relative to the effect of the
data. Instead, we hold out neurons rather than time bins. Accurate
network models play a crucial role in predicting held-out neurons'
activity, since the distribution of incident connections to the
held-out neuron are informed solely by the network model.

Formally, we estimate the probability of a held-out neuron's spike
train~$\bs_{n^*}=[s_{1,n^*}, \ldots, s_{T,n^*}]$, given the
observed spike trains. We integrate
over the latent variables and parameters underlying the
observed spike train, as well as those underlying the
new spike train, using Monte Carlo. 
Let~${\bZ = \{\{\bw_n, \ba_n, \nu_n, \bu_n, \bv_n\}_{n=1}^N, \btheta\}}$, and
let~${\bz_{n^*} = \{\nu_{n^*}, \bw_{n^*}, \ba_{n^*}, \bu_{n^*}, \bv_{n^*}\}}$.
Then,
\begin{align*}
  p(\bs_{n^*} \given \bS) \approx
  \int p(\bs_{n^*} \given \bz_{n^*}, \bS) \,
  p(\bz_{n^*} \given \bZ) \, &p(\bZ \given \bS) \,
  \mathrm{d} \bz_{n^*} \, \mathrm{d} \bZ 
  \approx
  \frac{1}{Q} \sum_{q=1}^Q p(\bs_{n^*} \given \bz_{n^*}^{(q)}, \bS), \\
  \bz_{n^*}^{(q)} \sim p(\bz_{n^*} \given \bZ^{(q)}),
  &\quad
  \bZ^{(q)} \sim p(\bZ \given \bS).
\end{align*}
The samples~$\{\bZ^{(q)}\}_{q=1}^Q$ are posterior samples generated
our MCMC algorithm given~$\bS$. While a proper Bayesian approach would
impute~$\bs_{n^*}$, for large~$N$ this approximation suffices. For
each sample, we draw a set of latent variables and connections for
neuron~$n^*$ given the parameters~$\bZ^{(q)}$. These,
combined with the spike train, enable us to compute the likelihood
of~$\bs_{n^*}$.

\section{Further MCMC details}

\paragraph{Gibbs sampling the parameters of the network model}

\begin{itemize}
\item \textit{Independent Model} Under an independent model, the
  neurons do not have latent variables so all we have to sample are
  the global parameters,~$\btheta$.  If the independent model applies
  to the adjacency matrix, then~$\btheta = \overline{\rho}$. The model
  is conjugate with a beta prior.  If the independent model applies to
  the weights, then~$\btheta = \{\overline{\bmu},
  \overline{\bSigma}\}$, and the model is conjugate with a normal
  inverse-Wishart prior.
  
  \item \textit{Stochastic Block Model (SBM) updates}:
    If a stochastic block model is used for either the adjacency matrix
    or the weights, then it is necessary to sample the class assignments
    from their conditional distribution. We iterate over each neuron and
    update its assignment given the rest by sampling from the conditional
    distribution. For example, if~$\bu_n$ governs a stochastic block model
    for the adjacency matrix, the conditional distribution of the label
    for neuron~$n$ is given by,
    \begin{align*}
      p(\bu_n = c \given \{\bu_{m \neq n}\}, \bA, \btheta)
      &\propto \pi_{c} \,
      \prod_{m=1}^N p(a_{m \to n} \given \rho_{c_m \to}) \,
                    p(a_{n \to m} \given \rho_{c \to c_m}),
    \end{align*}
    where~$\btheta = \{\bpi, \{\rho_{c \from c'}\} \}$. For stochastic block
    models of the weight matrix,~$\bW$, the conditional distribution
    depends on~$\bw_{n \to m}$ and~$\bw_{m \to n}$ instead.

    Given the class assignments and the network, the
    parameters~$\rho_{c \from c'}$,~$\bmu_{c \from c'}$,~$\bSigma_{c \from c'}$, and~$\bpi$ are easily updating
    according to their conditional distributions, assuming~$\bpi$
    and~$\rho_{c \to c'}$ are given conjugate Dirichlet and beta priors,
    respectively.

  \item \textit{Latent location updates}:
    We resample the locations using hybrid Monte Carlo (HMC) \citep{Neal10}.
    Since the latent variables are continuous and unconstrained,
    this method is quite effective.

    In addition to the locations, the latent distance model is parameterized
    by a location scale,~$\eta$. Given the locations and an inverse gamma
    prior, the inverse gamma conditional distribution can be computed in
    closed form.
    
    The remaining parameters include the log-odds,~$\gamma_0$, if the
    distance model applies to the adjacency matrix. This can be
    sampled alongside the locations with HMC.  For a latent distance
    model of weights, the baseline mean and
    variance,~$(\mu_0,\sigma^2)$, are conjugate with a normal
    inverse-gamma prior.
\end{itemize}

\paragraph{Observation parameter updates}
The observation parameter updates depend on the particular distribution.
Bernoulli observations have no parameters.
In the binomial model,~$\nu_n$ corresponds to the maximum number of
possible spikes --- this can often be set a priori, but it must
upper bound the maximum observed spike count.
For negative binomial spike counts, the shape parameter~$\nu_n$ can
be resampled as in~\citet{Zhou2012}. One possible extension is to
introduce a transition operator that switches between
binomial and negative binomial observations in order to truly capture
both over- and under-dispersion.

We implemented our code in Python using Cython and OMP to parallelize
the \polyagamma updates.  This is available at \url{https://github.com/slinderman/pyglm}.

\paragraph{Number of MCMC iterations}
For both the synthetic and retinal ganglion cell results presented in
the main paper, we ran our MCMC algorithm for 1000 iterations and used
the last 500 samples to approximate posterior expectations. These limits
were set based on monitoring the convergence of the joint log probability.
Even after convergence, the weighted adjacency matrix continues to vary
from sample to sample, reflecting genuine posterior uncertainty. In some
cases, the samples of the stochastic block model seem to get stuck in local
modes that are difficult to escape. This is a challenge with coclustering
models like these, and more sophisticated transition operators could be
considered, such as collapsing over block parameters in order to update
block assignments.

\end{document}